\documentclass[letterpaper]{article} 
\usepackage{aaai2026}  
\pdfoutput=1
\usepackage{times}  
\usepackage{helvet}  
\usepackage{courier}  
\usepackage[hyphens]{url}  
\usepackage{graphicx} 
\usepackage{natbib}  
\usepackage{caption} 
\usepackage{algorithm}
\usepackage{algorithmic}

\usepackage{newfloat}
\usepackage{listings}
\DeclareCaptionStyle{ruled}{labelfont=normalfont,labelsep=colon,strut=off} 
\floatstyle{ruled}
\newfloat{listing}{tb}{lst}{}
\floatname{listing}{Listing}
\title{Diffuse Thinking: Exploring Diffusion Language Models as Efficient Thought Proposers for Reasoning}
\author{
\textbf{Chenyang Shao}\equalcontrib \quad  \quad \textbf{Sijian Ren}\equalcontrib \\  \textbf{Fengli Xu} \quad  \quad \textbf{Yong Li} \\
Department of Electronic Engineering, BNRist, Tsinghua University, Beijing, China \\
\texttt{fenglixu@tsinghua.edu.cn}}

\usepackage[utf8]{inputenc} 
\usepackage{url}            
\usepackage{booktabs}       
\usepackage{arydshln}      
\usepackage{amsfonts}       
\usepackage{subfig}         
\usepackage{multirow}       
\usepackage{nicefrac}       
\usepackage{microtype}      
\usepackage[breakable]{tcolorbox}
\tcbuselibrary{skins}
\usepackage{amsmath}
\usepackage{graphicx} 
\usepackage{enumitem}
\usepackage{arydshln}
\usepackage{graphicx}
\usepackage{caption}
\nocopyright
\begin{document}
\maketitle

\begin{abstract}
In recent years, large language models (LLMs) have witnessed remarkable advancements, with the test-time scaling law consistently enhancing the reasoning capabilities. 
Through systematic evaluation and exploration of a diverse spectrum of intermediate ``thoughts'', LLMs demonstrate the potential to generate deliberate reasoning steps, thereby substantially enhancing reasoning accuracy.
However, LLMs' autoregressive generation paradigm results in reasoning performance scaling sub-optimally with test-time computation, often requiring excessive computational overhead to propose thoughts while yielding only marginal performance gains.
In contrast, diffusion language models (DLMs) can efficiently produce diverse samples through parallel denoising in a single forward pass, inspiring us to leverage them for proposing intermediate thoughts, thereby alleviating the computational burden associated with autoregressive generation while maintaining quality.
In this work, we propose an efficient collaborative reasoning framework, leveraging DLMs to generate candidate thoughts and LLMs to evaluate their quality.
Experiments across diverse benchmarks demonstrate that our framework achieves strong performance in complex reasoning tasks, offering a promising direction for future research. Our code is open-source at \url{https://anonymous.4open.science/r/Diffuse-Thinking-EC60}. 
\end{abstract}

\section{Introduction} \label{intro}

Since the explosive popularity of ChatGPT, transformer-based large language models (LLMs) have received unprecedented attention, becoming a research hotspot and sparking a wave of applications.
Influential models such as GPT-4o~\cite{GPT-4o}, Llama 3~\cite{LLaMA3}, and Deepseek-V3~\cite{liu2024deepseek}, which adopt a decoder-only transformer, as well as encoder-decoder models like ChatGLM~\cite{glm2024chatglm}, all adhere to the autoregressive generation paradigm. This paradigm necessitates a forward pass computation sequentially for each token decoded, leading to significant inference cost concerns that remain a primary challenge. 

Moreover, many existing methods to enhance the reasoning accuracy of LLMs, such as Chain-of-Thought (CoT)~\cite{wei2022chain} and Tree-of-Thought (ToT)~\cite{yao2023tree}, can be seen as attempting to scale greater token reasoning spaces to achieve higher reasoning accuracy. 
Recent work by OpenAI~\cite{openai2024learning} has brought test-time scaling into the spotlight, garnering significant attention within the academic and professional communities. This paradigm shifts the allocation of computational resources and time from the pretraining phase to the inference phase. Specifically, the reasoning models think more before the response, exploring a richer space of intermediate thoughts. By iteratively proposing diverse reasoning thoughts and coupling this process with systematic evaluation and optimization strategies, the models are able to improve their reasoning performance~\cite{Xu2025TowardLR,kong2025rethinking}.

However, this shift comes at a significant cost, as exemplified by OpenAI's o1 model~\cite{openai-o1}, which requires a hundredfold increase in token consumption to solve a single problem, thereby exacerbating the already substantial computational and monetary costs associated with large-scale generative models. 
Since generating tokens is more computationally demanding than processing inputs, the process of proposing thoughts constitutes the primary computational load, motivating us to explore whether there exists a method capable of efficiently generating reasoning proposals.


\begin{figure}[t]
\centering
\includegraphics[width=0.9\columnwidth]{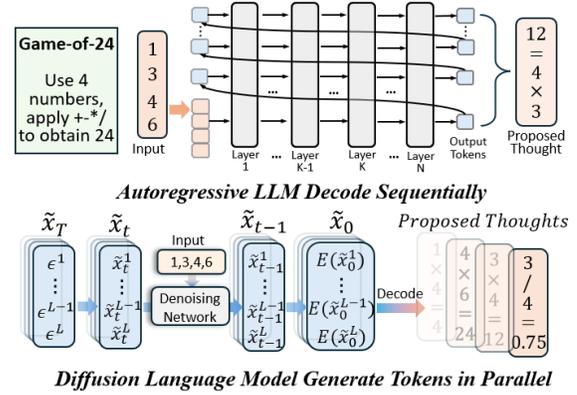} 
\caption{Comparison of generation in DLM and LLM: DLMs generate in parallel, producing multiple tokens simultaneously, while LLMs generate sequentially, one token at a time.}
\label{fig:compare}
\vspace{-6mm}
\end{figure}
The advancement of diffusion language models (DLMs) offers a promising direction, characterized by their efficient thought-proposing capabilities.
As shown in Figure \ref{fig:compare},
unlike LLMs which generate tokens sequentially and are constrained by their step-by-step, left-to-right nature, DLMs can produce multiple tokens simultaneously through reverse denoising.
DLMs parallelize the denoising process effectively, enabling them to simultaneously generate multiple proposals.
Moreover, while LLMs employ a uniform generation process for all tasks regardless of difficulty or quality requirements, DLMs can dynamically adjust their denoising steps, enabling a flexible trade-off between generation quality and speed tailored to the specific demands of the task.


These advantages position the DLM as a promising approach for proposing diverse proposals efficiently. On the other hand, while LLMs incur high output costs, they demonstrate relatively efficient input processing, as any length of text input can be processed and stored through a single forward pass. Combined with their strong semantic understanding capabilities, LLMs are well-suited to serve as evaluators, assessing and filtering intermediate reasoning proposals. This collaborative approach between the two models mitigates information loss in DLMs while enhancing the overall reasoning process. In contradistinction to speculative decoding, which performs token-level mimicry followed by per-token verification, the DLM-LLM collaboration is instantiated at the level of entire reasoning thoughts, engendering a synergistic co-reasoning paradigm.

However, directly employing existing DLMs as proposers for complex reasoning tasks presents several challenges. First, the collaboration mechanism between DLM and LLM is still unclear and the respective strengths of each model are not well understood. Second, while DLMs have certain advantages as proposers due to their ability to generate tokens in parallel, they are also prone to information loss during the generation process, as shown by information theory analysis. LLMs acting as an evaluator can mitigate the information loss that occurs when DLMs generate proposals. Informed by the aforementioned theoretical derivations, we propose a novel collaborative reasoning framework, ``Diffuse Thinking" (DT). Specifically, we demonstrate that the parallel generation capabilities of DLMs, when employed as proposers, combined with the evaluation and filtering capabilities of LLMs acting as evaluators, result in a more efficient and accurate reasoning process. To optimize this collaborative dynamic, we meticulously design the interaction between DLMs and LLMs.  
By focusing on single-step reasoning, DLMs can produce more relevant and accurate proposals. These proposals are subsequently evaluated by LLMs in their capacity as evaluators, allowing them to select the most promising solutions. We further create single-step data specifically for targeted fine-tuning of DLMs in their role as proposers, thereby enhancing their ability to generate high-quality reasoning thoughts.

We conduct extensive experiments on four logical and mathematical benchmarks. Across various benchmarks, the framework consistently outperforms baseline models in terms of accuracy while also achieving higher throughput. Specifically, the average accuracy and throughput of reasoning increase by 5\% and 10\%, respectively, compared to the most accurate baselines.
Our contributions can be summarized in three points: 
\begin{itemize}

\item We design a novel framework that combines DLM and LLM, where DLM is used to efficiently propose many proposals, which are subsequently evaluated by LLM. 


\item We theoretically reveal the sources of the advantages of our collaborative framework through information-theoretic and computational complexity analyses, demonstrating that it holds significant potential and is effective in practice. 

\item We validate the proposed framework across diverse logical and mathematical benchmarks, demonstrating that its synergistic design can achieve both higher reasoning accuracy and increased throughput simultaneously.
\end{itemize}

\section{Related works}

\subsection{Test-time scaling law in LLM reasoning}

The reasoning ability of large models is crucial for their application~\cite{Shao2024BeyondIG}. 
Pre-training once ruled LLM reasoning; recent work~\cite{openai2024learning,zhong2024evaluation} shows inference-time compute, yields large gains. 
This has led to the emergence of the \textit{Test-Time Scaling Law}. 
Through exploration and experimentation, researchers~\cite{wang2024openr, zhang2024llama, qin2024o1} have converged on a consistent implementation framework: 
Decompose problems, sample many thoughts, score a trained verifier (such as PRMs), and guide search (e.g., MCTS)~\cite{snell2024scaling,xie2024monte}.
Among these steps, the generation of a large number of thoughts constitutes the primary computational and time cost. This drives the need to develop an efficient method for proposing reasoning thoughts.
Existing methods primarily rely on the autoregressive generation capabilities of LLMs~\cite{Shao2025AgentStealthRL, Shao2025RouteandReasonSL}, which are computationally intensive, our work introduces a novel collaborative framework that leverages the parallel generation capabilities of Diffusion Language Models (DLMs) to efficiently propose reasoning thoughts. 



\subsection{Diffusion language models}
Diffusion models have garnered increasing attention in the field of text generation. To adapt the continuous diffusion latent space to the discrete nature of text tokens, two primary categories of DLMs have emerged:Discrete DLMs corrupt tokens via Markov transitions~\cite{zou2023survey, hoogeboom2021argmax, zheng2023reparameterized, loudiscrete}; continuous DLMs embed tokens and apply Gaussian noise. Both achieve strong results in MT, dialogue, summarization and controllable text~\cite{li2022diffusion}.
These approaches have been applied to tasks such as machine translation~\cite{nachmani2021zero}, controllable generation~\cite{han2022ssd, li2022diffusion}, dialogue~\cite{gong2022diffuseq}, and text summarization~\cite{zhang2023diffusum}, demonstrating their potential in natural language processing.
Beyond traditional NLP tasks, diffusion models are also being explored for complex reasoning. Pretrained models like Plaid~\cite{gulrajani2024likelihood} and SEDD~\cite{lou2023discrete} have shown text generation capabilities comparable to autoregressive models of similar scale. Efforts to tackle reasoning tasks include Diffusion-of-Thought~\cite{ye2024diffusion}, which integrates chain-of-thought reasoning into diffusion models, and work by~\cite{ye2024beyond}, which enhances training efficiency through token-level reweighting, achieving strong performance on tasks like Sudoku.  However, the latter remains limited to numeric tokens, lacking general text understanding. LLADA\cite{nie2025large} and DiffuLLaMA\cite{gong2024scaling} scale the DLMs to a size of 7-8B parameters. Mercury Coder\cite{mercury_tech_report}, a novel commercial-scale diffusion large language model for coding applications, achieves a remarkable generation speed of up to 1000 tokens per second, significantly outperforming traditional autoregressive models in terms of efficiency. These advancements highlight the potential of diffusion models as a versatile tool for language processing and reasoning tasks. 
Our work builds on these advancements by proposing a novel collaborative framework that combines the parallel generation capabilities of DLMs with the strong semantic evaluation capabilities of LLMs. 

\section{Preliminaries}

We begin by providing a formal description of the generation processes for DLMs and LLMs, establishing a foundation for the subsequent complexity analysis.

\subsection{Diffusion language models}

\paragraph{Forward diffusion process}  
Given an input token sequence \(x = [x^{(1)}, \ldots, x^{(L)}]\), mapped to embeddings \(\tilde{x} = [\tilde{x}^{(1)}, \ldots, \tilde{x}^{(L)}]\), the forward noising process at step \(t\) is defined as:
\[
\tilde{x}_t = \sqrt{\bar{\alpha}_t} \tilde{x}_0 + \sqrt{1 - \bar{\alpha}_t} \epsilon,
\]
where \(\bar{\alpha}_t = \prod_{i=1}^{t} \alpha_i\), \(\epsilon \sim \mathcal{N}(0, I)\), and \(\alpha_t = 1 - \beta_t\) controls the noise schedule.

\paragraph{Reverse generative process}  
The reverse denoising process is modeled as a Markov chain. At each step \(t\), the model predicts the noise \(\mathbf{\epsilon}_\theta(\tilde{x}_t, t)\) and computes the denoised mean:
\[
\mu_\theta(\tilde{x}_t, t) = \frac{1}{\sqrt{\alpha_t}} \left( \tilde{x}_t - \frac{1 - \alpha_t}{\sqrt{1 - \bar{\alpha}_t}} \mathbf{\epsilon}_\theta(\tilde{x}_t, t) \right).
\]
The reverse process iteratively samples \(\tilde{x}_{t-1}\) from \(p_\theta(\tilde{x}_{t-1} | \tilde{x}_t)\) until \(\tilde{x}_0\) is obtained.

\subsection{Autoregressive large language models}

\paragraph{Inference process}  
For an input sequence \(x^{<t} = [x^1, x^2, \ldots, x^{t-1}]\), the model predicts the next token \(x^t\) by computing:
\[
p(x^t | x^{<t}; \theta) = \text{softmax}(W h^{<t} + b),
\]
where \(h^{<t} = \text{Transformer}(x^{<t})\) is the hidden state produced by a Transformer decoder or encoder-decoder, depending on the specific architecture, and \(W, b\) are model parameters. The joint probability of the sequence is:
\[
p(x) = \prod_{t=1}^{n} p(x^t | x^{<t}; \theta).
\]
Text generation proceeds iteratively, sampling tokens until an end token or maximum length is reached.

\subsection{Symbolic definition of network layer architecture}

As the subsequent discussion encompasses the derivation of the model's computational complexity and time complexity, which will involve specific network layers, we establish a unified notation as follows:
Assume the input text token length for the reasoning task is \( L_{in} \), and the output length of the language model is \( L_{out} \). For DLM, \( L_{in} = L_{out} \). We require each model to generate \( K \) proposals. Moreover, we denote the transformer model dimension as $D$, embedding dimension as $E$, number of attention heads as $H$, number of Transformer blocks as $N$, vocabulary size as $V$, and number of denoising steps as $T$.

\section{Methodology}

\begin{figure*}
    \centering
    \includegraphics[width=0.8\linewidth]{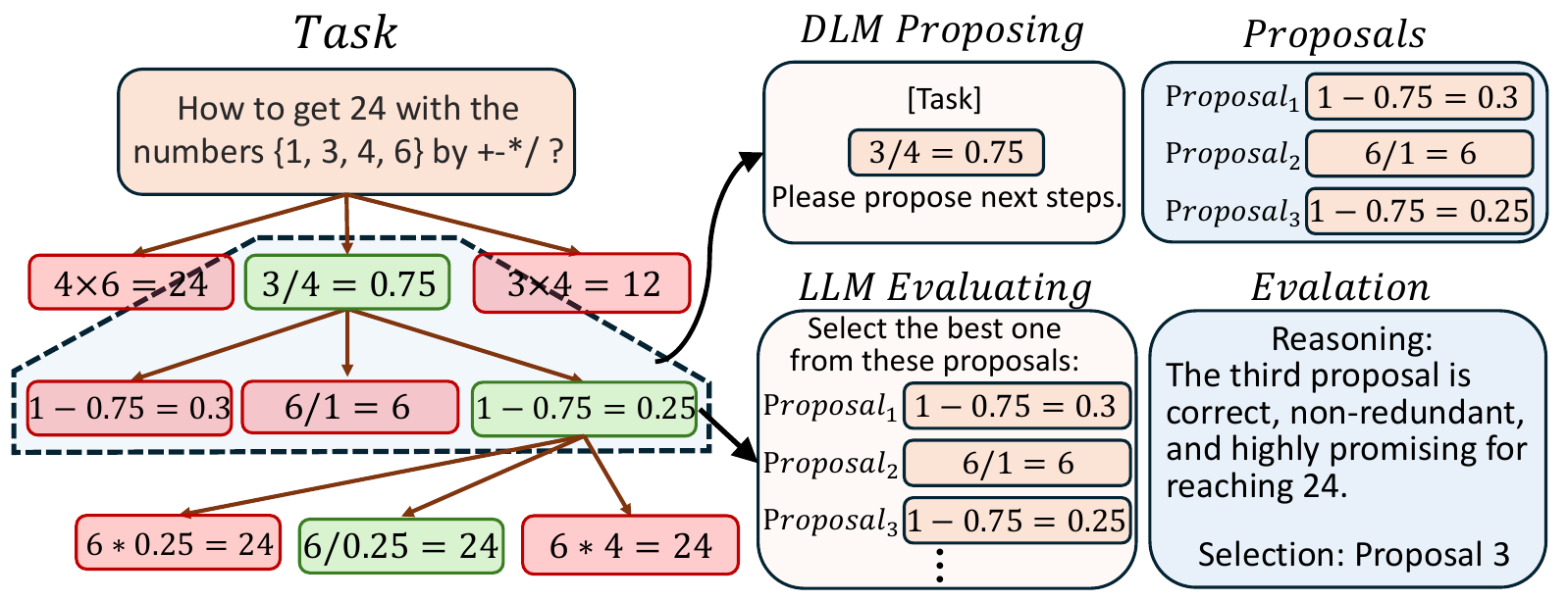}\\[-1ex]
    \caption{Our proposed framework. The DLM efficiently generates multiple reasoning thoughts in parallel, exemplified by the Game of 24. These thoughts are then evaluated and selected by the LLM, which selects the most promising proposal.}
    \label{fig:enter-label}
    \vspace{-3ex}  
\end{figure*}

In this section, we introduce the proposed collaborative framework that integrates the DLM and LLM for problem-solving. The DLM leverages its strength in efficient parallel generation by progressively denoising candidate proposals, while the LLM excels in comprehensive semantic understanding, enabling it to evaluate these proposals and provide well-reasoned selections.
The overall framework follows an iterative stepwise optimization paradigm: the reasoning problem is naturally decomposed into several subproblems. For each subproblem, the DLM explores extensively by generating multiple candidate thoughts through diverse inference and solution attempts. The LLM then evaluates and filters these thoughts, selecting the most promising ones to be incorporated into the solution. This process continues iteratively, with the framework repeatedly proposing and evaluating  until getting the final answer.

\subsection{DLM as thought proposer}
\paragraph{Computational complexity derivation of DLM
} For DLM, since the input and output lengths are fixed and equal, the computational complexity is primarily determined by the denoising transformer network layers and the number of denoising steps. For $T$ denoising steps, the total FLOPs for the DLM can be expressed as:
\begin{equation}
\begin{aligned} 
F_{\text{DLM}} &= T \cdot \left( 2\cdot N \cdot \left( 12 \cdot L \cdot D^2 + 2 \cdot L^2 \cdot \frac{D}{H} \right) \right.\\ 
&\quad \left. +2\cdot L \cdot D \cdot E + 2\cdot L \cdot (D + 2 \cdot E) \cdot V \right) \\
&= \mathcal{O} \left( T\cdot L^2\right)
\end{aligned}
\end{equation}
For a more detailed derivation, please refer to Appendix. Considering the parallel generation of \(K\) samples, the overall computational cost becomes:
\begin{equation}
    F_{	\text{DLM (K)}} = K\cdot F_{	\text{DLM}}
\end{equation}

\paragraph{Parallelization efficiency}
However, in actual computations, when generating \( K \) proposals, the generation time does not increase linearly by a factor of \( K \).
For DLMs, the input and output lengths of multiple samples are the same, and the number of denoising steps remains identical. This setting enables fully parallel denoising. We define a parallel efficiency factor \( \beta \) to model the efficiency gains from parallelization, causing the time complexity to scale sublinearly:
\begin{equation}
\label{eq:dlm}
    T_{	\text{DLM}} = \mathcal{O}\left( K_{DLM}^{\beta} \cdot L^2 \cdot T \right), \quad 0 \leq \beta \leq 1
\end{equation}

\paragraph{Parallel-time complexity.}  
All matrix multiplications inside self-attention and MLP are parallelizable over the sequence dimension~$L$, since the DLM processes the \emph{entire sequence simultaneously} in forward passes.
Assuming ideal hardware that provides $O(L)$ parallel lanes, the \emph{wall-clock latency} per denoising step collapses to the longest dependency path, yielding
\begin{equation}
T_{\text{DLM}}^{\text{par}} = O\!\left(\frac{K_{DLM}^{\beta} \cdot L^2 \cdot T}{L} \right)=O(K_{DLM}^{\beta} \cdot L \cdot T).
\end{equation}
Thus, under sufficient parallelism, the latency is \emph{linear in}~$T$ and \emph{linear in}~$L$, while the total FLOPs still scale as~$O(L^{2}T)$.

\paragraph{Computational complexity derivation of LLM
}In contrast to DLM, the computational cost of LLMs is determined by both their backbone architecture and the variable length of output text. We denote the per-head dimension as:
\(
d_h = \frac{D}{H}.
\)
Considering KV-cache, the computational complexity of LLM can be expressed as:
\begin{equation}
\begin{aligned}
\label{eq:llm0}
    F_{\text{total}} = 
    &\quad (2\cdot D \cdot V +24 \cdot N \cdot D^2) \cdot L_{out}\\\notag
    &\quad +(2+24 \cdot N \cdot D)\cdot D \cdot L_{in}  \\\notag
    &\quad + 4 \cdot {N} \cdot \frac{D}{H} \cdot\left( L_{in}^2+L_{out}^2+L_{in}\cdot L_{out}\right)).
\end{aligned}
\end{equation}
For a more detailed derivation, please refer to Appendix.

When generating \( K \) samples, the complexity scales by a factor of \( K \) and is further affected by the length of the longest generated sequence:
\begin{equation}
\label{eq:llm}
\begin{split}
    T_{\text{LLM}} 
    &= \mathcal{O}\!\Bigl(K_{\text{LLM}}\bigl(L_{\text{out,max}}^{2}
      +L_{\text{in}}^{2}
      +L_{\text{in}}\,L_{\text{out}}\bigr)\Bigr) \\[1mm]
    &=\mathcal{O}\bigl(K_{\text{LLM}}\,L^{2}\bigr)
\end{split}
\end{equation}

\paragraph{Sequential-time complexity.}
Unlike diffusion language models, autoregressive language models generate tokens one by one; each new token depends on all previously generated tokens. Consequently, self-attention and MLP layers cannot be parallelized across the sequence dimension~$L_{\text{out}}$, even under ideal hardware.

We also computationally demonstrate that $K_{DLM}$ can reach 16 times as much as $K_{LLM}$ for the same model architecture, computational resources, parameter sizes, and input/output lengths while best fitting our experimental setup. The above are all derived results considering KV-cache for LLM. If the KV-cache is not considered, the computational cost of the LLM increases rapidly as the sequence length increases, due to the cubic dependence on $L_{\text{out}}$, while the computational cost of the diffusion model remains low. For a more detailed derivation, please refer to Appendix.

The comparison of DLM and LLM highlights the significant advantage of DLM's parallel generation, which inspires us to fully leverage the parallelism of DLMs to replace LLMs for proposing a number of thoughts, thereby greatly improving inference efficiency. 
Assuming that the DLM generates $M$ thoughts in parallel, then the denoising process can be defined as follows:
\begin{equation}
    \mathcal{P} = \{\text{Denoised}_{(j)}(\tilde{x}_0) \mid j = 1, \ldots, M\}    
\end{equation}

``Denoised'' represents the process in which DLM gradually denoises from noise and generates final outputs. The equation as a whole indicates that DLM generates M thoughts, which are placed in a set $\mathcal{P}$ for subsequent derivation.

\subsection{LLM as thought evaluator}

The underlying mechanisms of inference in LLMs and DLMs differ fundamentally. For LLMs, with the support of KV cache, the process of handling inputs and generating tokens is separated into two phases. In the first phase, known as prefill, LLM processes all input tokens in parallel to initialize the KV cache, performing a single forward pass to output the first token. The second phase is decoding, where the model generates subsequent tokens one by one, requiring a forward pass for each generated token. The existence of these two phases explains why input tokens are cheaper than output tokens in current LLMs. However, for DLMs, handling inputs and generating outputs cannot be separated, and there is no advantage in processing inputs more efficiently than generating outputs. This indicates that LLMs exhibit high efficiency in processing input. Regardless of the input text length, an LLM only requires a single forward pass to encode the entire input into the initial state of the attention memory. This advantage is not present in DLMs, which inspires us to leverage LLMs as evaluators to assess and filter a large number of proposed thoughts. Moreover, LLMs inherently possess strong general semantic understanding capabilities, enabling them to identify the most promising and correct proposal from a set of candidates.

Specifically, once the DLM generates a set of candidate solutions, LLM can be leveraged to evaluate and select the most appropriate one, taking into account the context provided by the problem and the proposals. The evaluation process can be formulated as follows:

The LLM receives a prompt that includes the problem \(q\), the candidate solutions \(\mathcal{P}\), and relevant guidance. The model then predicts both the most suitable solution index \(i^*\) and the corresponding reasoning \(r\) in a unified process by computing the joint probability distribution:
\begin{equation}
    p_\text{LLM}(i^*, r | q, \mathcal{P}; \theta)    
\end{equation}
Here, \(i^*\) represents the chosen solution index, and \(r\) is the reasoning behind the choice. This formulation encapsulates both solution selection and reasoning generation in a single model evaluation.
This formulation enables the LLM to provide both a solution and an explanation in one inference pass, significantly reducing computational overhead compared to evaluating each candidate solution individually.

\subsection{DLM-LLM collaborative reasoning framework}
\newtheorem{theorem}{Theorem}
\newtheorem{lemma}{Lemma}
\newtheorem{definition}{Definition}
\newtheorem{proposition}{Proposition}
\newtheorem{corollary}{Corollary}

\newcommand{\MI}{I} 
\newcommand{\Ent}{H} 
\newcommand{\Prob}{P} 
\newcommand{\Expect}{\mathbb{E}} 
Finally, we integrate the powerful capability of the DLM to efficiently generate diverse solutions with the reasoning and comprehension abilities of the LLM. By leveraging an iterative propose-evaluate tree-structured reasoning framework, we construct a highly efficient and comprehensive inference system. The overall process proceeds as follows: First, the diffusion model generates a set of candidate solutions \(\mathcal{P}\) by reversing the noise from the initial noisy states \(\tilde{x}_T^{(j)}\) for each \(j = 1, \ldots, M\). After generating the candidate solutions, the problem input \(q\) is concatenated with \(\mathcal{P}\) to form a prompt for the LLM.

The LLM then evaluates all the candidates in one inference step, predicting the most suitable solution index \(i^*\) and providing the corresponding reasoning \(r\). The selected solution is the one with the highest probability:
\begin{equation}
    i^* = \arg\max_{i} p_\text{LLM}(i | q, \mathcal{P}; \theta)
\end{equation}
Finally, the selected solution \(x^*\) is obtained as \(x^* = x_{i^*}\), and the reasoning \(r\) provides an explanation for the selection.

We present a theoretical analysis that elucidates the fundamental limits of error in diffusion language models from an information-theoretic perspective. Specifically, the per-step information loss \(\Delta \MI_t\) is defined as the difference between the model's estimated mutual information and the true mutual information:

\begin{equation}
\Delta \MI_t = \MI_{\text{indep}}(X_t | X_{t-1}) - \MI(X_t | X_{t-1}) \ge 0.
\end{equation}

This loss quantifies the information about the dependencies between tokens \(x_i^t\) within step \(t\) (given \(X_{t-1}\)) that is omitted by the model. 

Let $\MI_{\text{ideal}}(X^*; X_0)$ be the mutual information achievable by an ideal T-step diffusion process (without the independence assumption), and $\MI_{\text{indep}}(X^*; X_0)$ be the mutual information achieved by the model using the independence assumption. 

The total accumulated information loss is:
\begin{equation} \text{TotalLoss}(T, L) = \MI_{\text{ideal}}(X^*; X_0) - \MI_{\text{indep}}(X^*; X_0) \ge 0 \end{equation}

This total loss reflects the cumulative impact of the per-step losses \(\Delta \MI_t\) across all denoising steps. 
We derive the following inequality using Fano's inequality\cite{scarlett2019introductoryguidefanosinequality}:
\begin{multline}
\Ent_{\text{ideal}}(X^{*} \mid X_{0})
+ \text{TotalLoss}(T, L) \\
\le \Ent\!\bigl(E_{\text{Diff}}(T, L)\bigr)
+ E_{\text{Diff}}(T, L)\log\bigl(|\mathcal{X}| - 1\bigr).
\end{multline}
$\text{TotalLoss}(T, L)$ denotes the total accumulated information loss, while $\Ent_{\text{ideal}}(X^* | X_0)$ represents the conditional entropy of the ideal denoising process. $E_{Diff}(T, L) = \Prob(X_0 \ne X^*)$ denotes the final error probability, and $\mathcal{X}$ is a finite set of possible values for $X^*$. A detailed derivation of this inequality is provided in Appendix. This indicates that the lower bound of error in the diffusion language models is affected by the loss of information in the parallel prediction tokens, which is expected to increase as $L$ increases. By integrating the diverse solution generation capabilities of diffusion models with the global reasoning capabilities of large language models (LLMs), the proposed framework systematically optimizes the solution for each step, narrowing the gap with the correct answer for that step. The LLM not only selects the optimal solution but also ensures transparency by providing a detailed explanation of its choice. 

\subsection{Learning to propose thoughts}
\label{methods:training}

To further elicit the intrinsic capability of DLMs for continuous and incremental problem-solving through training, we meticulously design the construction of the training data. Specifically, for each reasoning task, we perform task decomposition, breaking down the complete reasoning process into sub-tasks that are more manageable and relatively easier to solve. 
Each sub-task corresponds to a specific and practical reasoning thought, such as computation, concatenation, transformation, and so forth.
Each ground truth thought is obtained through a \textit{search}-based approach, where we explore and traverse the entire solution space. Ultimately, we filter and retain the most accurate and promising thoughts to be included in the dataset.
Each thought constitutes a part of the solution, effectively advancing the problem-solving process.
For each training instance, the input consists of the problem description of the reasoning task along with the preceding reasoning thoughts, while the output is the next reasoning thought.

This dataset construction approach encourages our DLM to spontaneously adopt step-by-step reasoning to tackle problems and propose the next-step reasoning thoughts. The DLM can perform parallel denoising on a batch to effectively generate multiple next-step thoughts, forming a set of candidate proposals. These proposals are then evaluated and filtered, with one or several selected to proceed, enabling the iterative process of proposing and filtering until the problem is solved.

\section{Experiments}

\begin{table*}[t]
\centering
\caption{Performance of our proposed framework(\textbf{DT}). }
\label{tbl:mainResult}
\renewcommand\arraystretch{1.2}
\setlength{\tabcolsep}{0.5mm}
\resizebox{0.85\textwidth}{!}
{

\begin{tabular}{c|cc|cc|cc|cc}
\multirow{2}{*}{\begin{tabular}[c]{@{}c@{}}\textbf{Solutions}\\\textbf{(Proposer + Evaluator)}\end{tabular}} & \multicolumn{2}{c|}{\textbf{Game of 24}} & \multicolumn{2}{c|}{\textbf{Trip-Planning}} & \multicolumn{2}{c|}{\textbf{GPQA}} & \multicolumn{2}{c}{\textbf{ARC-C}}  \\ 
\cline{2-9}
                                                                                                             & Acc    &   ThroughtPut                                             & Acc&   ThroughtPut & Acc    &  ThroughtPut& Acc    &   ThroughtPut                                                                                      \\ 
\hline

Llama3-8B + Llama3.3-70B & 0.12 & 0.30 & 0.04 & 0.23 & 0.32 & 0.13 & 0.80 & 0.49  \\
Mistral-7B + Llama3.3-70B & 0.08 & 0.41 & 0.07 & 0.18 & 0.33 & 0.19 & 0.81 & 0.57  \\
Deepseek-7B + Llama3.3-70B & 0.12 & 0.36 & 0.11 & 0.25 & 0.33 & 0.38 & 0.82 & 0.53  \\
\cdashline{1-9}
\textbf{DT(LLADA-8B + Llama3.3-70B)} & 0.12 & \textbf{0.48} & 0.16 & 0.35 & 0.33 & \textbf{0.56} & 0.82 & \textbf{1.05}  \\
\textbf{DT(Dream-7B + Llama3.3-70B)} & \textbf{0.13} & 0.45 & \textbf{0.29} & \textbf{0.40} & \textbf{0.37} & 0.46 & \textbf{0.83} & 0.59 
\end{tabular}
}
\vspace{-5mm}
\end{table*}

We test the comprehensive reasoning performance and reasoning efficiency of our proposed framework on four challenging logical benchmarks, and compared them with baselines.

\subsection{Experimental setup}

\textbf{Benchmarks:}
We select the Game-of-24, GPQA, ARC-C and Trip-Planning problems as our evaluation benchmarks. 
Game-of-24 requires using four given numbers to compute the number 24 through addition, subtraction, multiplication, and division, assessing both fundamental arithmetic skills and the ability to maintain a global logical perspective throughout the reasoning process. 
GPQA presents a PhD-level task that demands identifying the single correct answer among four choices in biology, physics, or chemistry, foregrounding retrieval of rare factual knowledge, chaining of multi-step scientific reasoning, and resistance to subtle, expert-designed distractors.
ARC-C tasks the solver with electing the single valid choice from four elementary-level science options, spotlighting common-sense induction, unseen domain transfer over distant concepts, and immunity to surface-level semantic mirages.
Trip-Planning entails composing a uniquely-valid multi-city itinerary under deterministic flight-connectivity and day-level duration constraints, foregrounding implicit constraint propagation, cross-domain temporal reasoning.

For the Game-of-24 , we exhaustively enumerated number combinations and employed a backtracking approach to solve it. We considered all non-repetitive, diverse solutions, resulting in a dataset of 168,046 24-point problems, with the maximum value for the four numbers limited to 30.
For GPQA benchmark, we used the original dataset from~\cite{rein2024gpqa}.
For ARC-C, we used the the challenge set from the original dataset from~\cite{clark2018think}.
For trip planning, we used the 3-city subset of the original dataset from~\cite{zheng2024natural}.
All the datasets used and created in this paper will be open-sourced at the project link.

\subsubsection{Metrics}
\textbf{\textit{Acc}} With the exception of the Game-of-24, all other benchmarks possess a unique solution. We compare the generated results against the ground truth to compute the accuracy of reasoning. For the Game-of-24, in addition to verifying whether the final result equals 24, we also scrutinize the correctness of the intermediate steps, such as whether all four numbers have been utilized adequately and whether each computational step is free from errors.

\textbf{\textit{Throughput}} Furthermore, adhering to a broad experimental setup, we employ \textit{throughput} as a metric to measure the number of samples processed per minute (it/min) during inference with a batch size of one, thereby evaluating the time efficiency of the framework in problem-solving.

\textbf{\textit{Time}} To intuitively demonstrate efficiency, we adopt \textit{time} as the measurement standard in single step evaluation, measuring the average time (in seconds) used during inference.

\textbf{\textit{Pass@5}} To examine the quality of the thoughts, we measure the probability that at least one of the five proposals is correct for a single sub-question.

\textbf{Baselines}

Given that the parameter count of the DLM model falls within the range of 7 and 8 billion, to ensure a fair comparison, we restrict our consideration to LLM models with parameters close to 8 billion. Ultimately, we select the recently released and widely recognized models: Llama3-8B~\cite{LLaMA3}, Mistral-7B~\cite{jiang2023mistral7b} and Deepseek-7B~\cite{bi2024deepseek} as proposers. We configure the evaluation model, Llama3-70B~\cite{LLaMA3}, to assess the thoughts generated by each proposer.

\textbf{Implementation details}
\label{experiment implementation details}
In inference, the number of sampling steps \( T \) is dynamic. This demonstrates the flexibility of Diffusion LM and improves the efficiency of proposal generation. We set \( T = 8 \) for Game-of-24, \( T = 64 \) for GPQA ,Trip-Planning and ARC-C. In the sections pertaining to Main results and Quality of proposals, our DLM leverages two open-source DLMs, specifically Dream~\cite{dream2025} and LLADA~\cite{nie2025large}. 
We finetune our model on one NVIDIA A100-80G GPUs with a batch size of 2,  and the sampling process is also conducted on a single NVIDIA A100-80G GPU.
Our DLM is finetuned based on the pretrained discrete diffusion language model, LLADA~\cite{nie2025large}. 
During finetuning, we set the learning rate to 1e-5 and the maximum sequence length to 4096.

\vspace{-2mm}
\subsection{Main results}
\vspace{-2mm}
\label{main results}
The comparison between our proposed method and the baseline results is presented in Table \ref{tbl:mainResult}. In Table \ref{tbl:mainResult}, DLM has not been fine-tuned.

Under the same evaluator, our framework achieves comparable accuracy to the baseline model while attaining a higher throughput, demonstrating improved generation efficiency. Notably, on the Trip-Planning benchmarks, our framework exhibits a clear advantage in accuracy, highlighting the capability of DLMs as proposers to generate high-quality reasoning thoughts on par with LLMs.

\begin{table}[t]
\centering
\caption{Quality of single step proposals}
\label{tbl:propose}
\renewcommand\arraystretch{1}
\setlength{\tabcolsep}{0.5mm}
\resizebox{\linewidth}{!}
{

\begin{tabular}{c|cc|cc|cc|cc}
\multirow{2}{*}{\begin{tabular}[c]{@{}c@{}}\textbf{Models}\end{tabular}} & \multicolumn{2}{c|}{\textbf{\textbf{Game of 24}}} & \multicolumn{2}{c|}{\textbf{Trip-Planning}} & \multicolumn{2}{c|}{\begin{tabular}[c]{@{}c@{}}\textbf{GPQA}\end{tabular}}& \multicolumn{2}{c}{\begin{tabular}[c]{@{}c@{}}{\textbf{ARC-C}}\end{tabular}}   \\ 
\cline{2-9}
                                                                                                             & pass@5    & Time                              & pass@5    & Time               & pass@5    & Time & pass@5    & Time                                                                   \\ 
\hline
Llama3-8B & 0.15 & 6.43 & 0.33 & 42.12 & 0.55 & 78.28& 0.88 & 28.32   \\
Mistral-7B & 0.065 & 4.76 & 0.33 & 40.62 & 0.25 & 49.94& 0.83 & 18.88  \\
Deepseek-7B & 0.125 & 18.06 & 0.36 & 38.94 & 0.33 & 55.71& 0.85 & 15.21   \\
\textbf{LLaDA-8B} & 0.165 & \textbf{2.16} & \textbf{0.60} & \textbf{23.86} & \textbf{0.57} & \textbf{10.19}& 0.87 & \textbf{13.25}  \\
\textbf{Dream-7B} & \textbf{0.185} & 3.97 & 0.47 & 29.91& 0.56 & 28.48& \textbf{0.89} & 18.31  \\
\end{tabular}
}

\vspace{-5mm}
\end{table}

\begin{figure*}[htbp]
    \centering
    \subfloat[Game of 24 \label{fig:scaling_accuracy_benchmark1}]{%
        \includegraphics[width=0.43\linewidth]{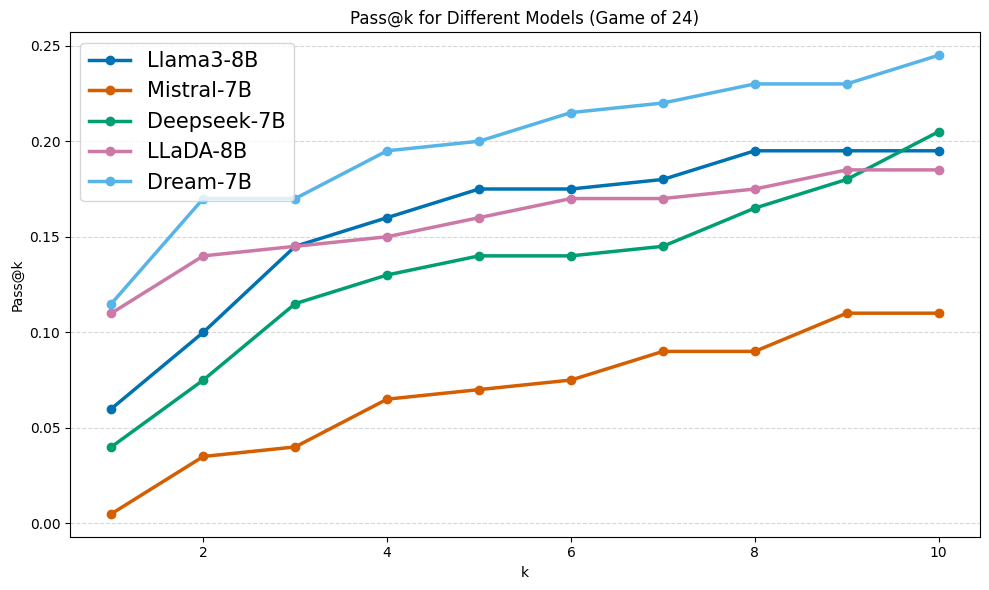}%
    }
    \hspace{0.03\linewidth}
    \subfloat[Trip planning \label{fig:scaling_accuracy_benchmark2}]{%
        \includegraphics[width=0.43\linewidth]{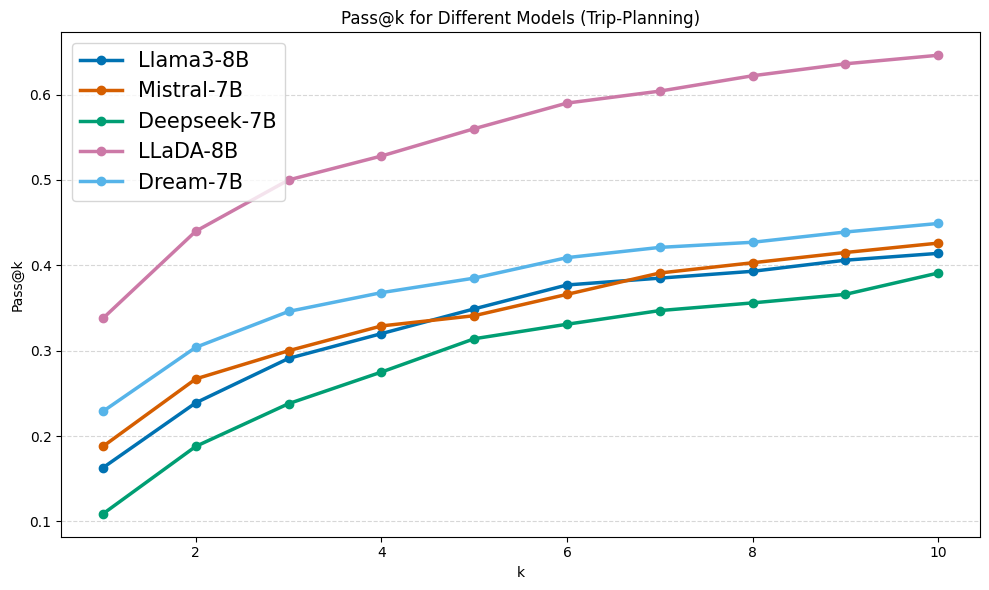}%
    }

    \vspace{-0.3cm}

    \subfloat[GPQA \label{fig:scaling_accuracy_benchmark3}]{%
        \includegraphics[width=0.43\linewidth]{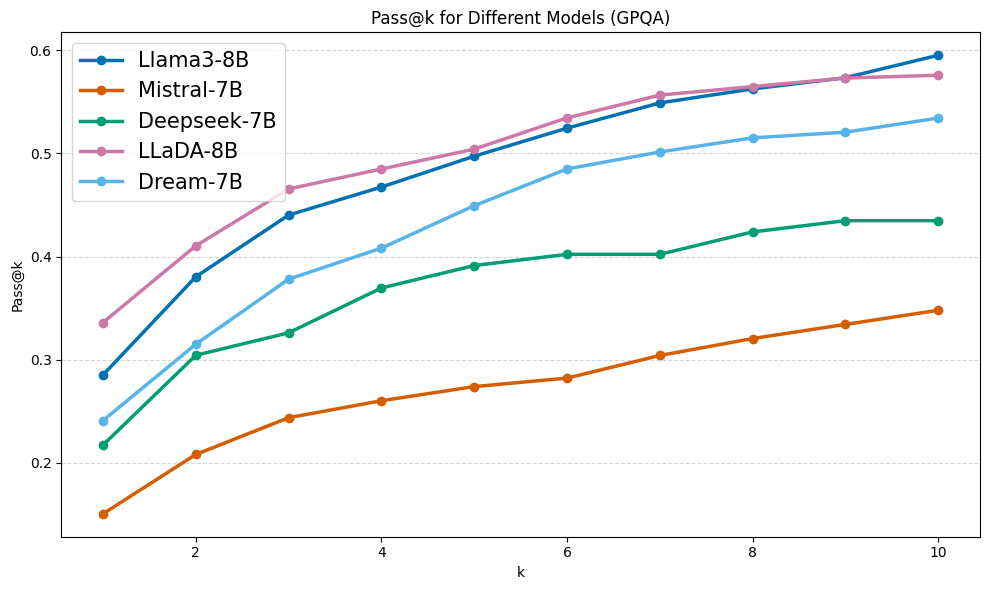}%
    }
    \hspace{0.03\linewidth}
    \subfloat[ARC-C \label{fig:scaling_accuracy_benchmark4}]{%
        \includegraphics[width=0.43\linewidth]{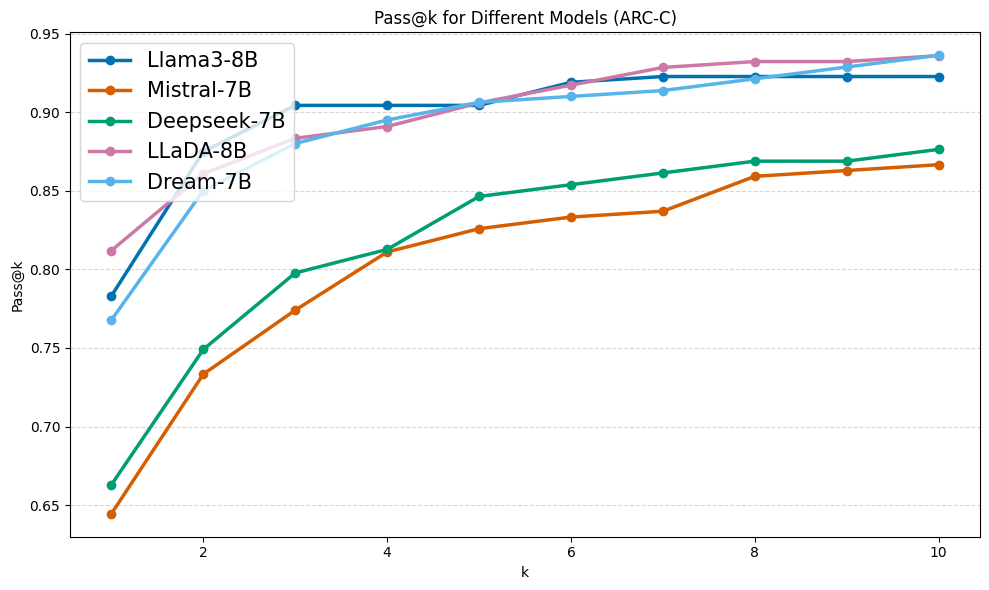}%
    }
    \vspace{-0.3cm}
    \caption{Proposal accuracy scaling with varying proposal quantities (1--10) across different benchmarks.}
    \label{fig:scaling_accuracy}
    \vspace{-0.3cm}
\end{figure*}
\vspace{-1mm}
\subsection{Quality of proposals}
Furthermore, to examine the quality of the thoughts proposed by the DLM without finetuning, we omitted the LLM evaluator and independently assessed the quality of the thoughts generated by several proposers. The results in Table~\ref{tbl:propose} clearly demonstrate that on different benchmarks, the quality of reasoning generated by DLM is better than that of the baseline models or at least comparable to the baseline models.
The average accuracy advantage exceeds 10\%, and the generation efficiency is also notably better, with an average throughput advantage surpassing 15\%. This fully corroborates our previous complexity analysis of the DLM and LLM models.

\vspace{-2mm}
\subsection{Scaling of Proposals}
\label{subsec:scaling_of_proposals}

In this section, we conduct an in-depth analysis of the \textit{Test-time Scaling} properties of the DLMs, focusing on how reasoning performance scales with the number of generated proposals. Specifically, we analyze the trade-off between reasoning accuracy and compute time by examining how varying the number of proposed thoughts impacts the single-step performance. We randomly select 100 tasks and conducted small-scale experiments by incrementally increasing the number of proposals from 1 to 10.

The results, presented in Figure~\ref{fig:scaling_accuracy}, show that increasing the number of proposals continuously enhances reasoning accuracy. However, after reaching 8 the rate of improvement gradually levels off and approaches a deterministic maximum.
This indicates that there exists a threshold in the trade-off between benefit and efficiency. 
Beyond this threshold, increasing the number of proposals results in diminishing improvements in inference accuracy.
Through this analysis, we aim to identify the optimal operating points that balance computational resources with performance gains.

\vspace{-3mm}
\subsection{Learning to propose thoughts}
\vspace{-1mm}
\begin{figure}[htbp]
\centering
\includegraphics[width=\linewidth]{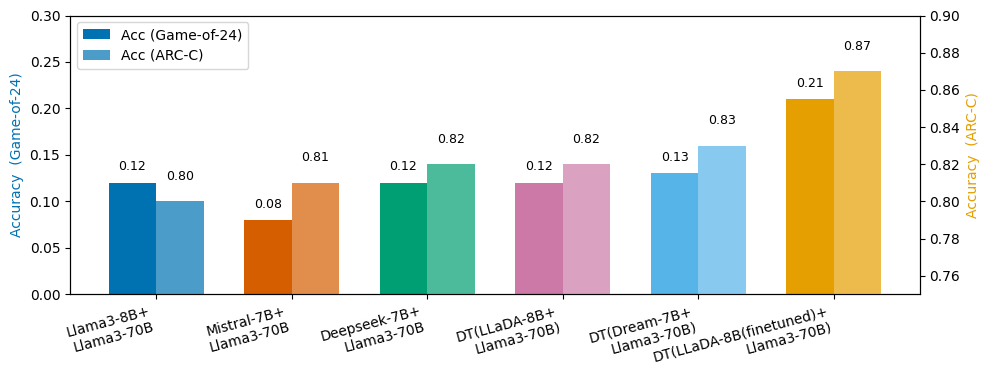}
\vspace{-6mm}
\caption{Performance improvements on different benchmarks after fine-tuning.}
\label{fig:game_of_24_improvement}
\vspace{-3mm}
\end{figure}
In this subsection, we demonstrate the significant performance improvements achieved by fine-tuning our model on two benchmarks: ARC-C and Game of 24. Figures \ref{fig:game_of_24_improvement} illustrates the performance gains on the ARC-C and Game of 24 datasets, respectively. As shown in the figures, the model's performance on the ARC-C dataset increased from 13\% to 21\% after fine-tuning, while on the Game of 24 dataset, the performance improved from 83\% to 87\%. These results underscore the effectiveness of fine-tuning DLM as an efficient proposer in enhancing our framework’s reasoning and problem-solving capabilities.

\vspace{-2mm}
\section{Conclusion}

This paper proposes integrating DLMs with LLMs to tackle complex logical and mathematical reasoning. By using DLMs to propose effective subproblem solutions and leveraging LLMs for evaluation and selection, the framework improves reasoning accuracy and efficiency.

\newpage
\appendix

\section{Detailed FLOPs calculation for diffusion language model} 
\label{appendix:dlm-flops}

In this section, we will provide a detailed derivation and computation of the forward inference computational cost (in FLOPs) for diffusion models that use a transformer decoder as their backbone.

\subsection{Symbol definitions and assumptions}
The following symbols are used: sequence length $L = 4096$, model dimension $D = 4096$, embedding dimension $E = 4096$, number of attention heads $H = 32$, number of Transformer blocks $N = 32$, vocabulary size $V =  126,464$, and number of denoising steps $T$ (variable). Moreover, $F_{\text{SA}}$ represents FLOPs for a single self-attention operation. $F_{\text{MLP}}$ represents FLOPs for a single feedforward network. $F_{\text{others}}$ represents FLOPs for embedding and other operations.


\subsection{FLOPs for self-attention and feedforward}
The self-attention (SA) mechanism includes the following components:
1. Generating queries, keys, and values: $F_{\text{QKV}} = 2 \cdot 3\cdot L \cdot D^2$.
2. Computing attention scores and normalization: $F_{\text{attention\_scores}} = 2 \cdot L^2 \cdot D/H$.
3. Weighted sum and output projection: $F_{\text{weighted\_sum}} + F_{\text{out\_projection}} = 2 \cdot L^2 \cdot D + 2 \cdot L \cdot D^2$.

Thus, the total FLOPs for SA is:
\begin{equation}
  F_{\text{SA}} = 8 \cdot L \cdot D^2 + 4 \cdot L^2 \cdot D/H
\end{equation}
The feedforward network (MLP) consists of up-projection and down-projection, each contributing $8 \cdot L \cdot D^2$, resulting in:
\begin{equation}
  F_{\text{MLP}} = 16 \cdot L \cdot D^2
\end{equation}
For one Transformer block, the total FLOPs is:
\begin{equation}
  F_{\text{block}} = F_{\text{SA}} + F_{\text{MLP}} = 24 \cdot L \cdot D^2 + 4 \cdot L^2 \cdot D/H
\end{equation}

\subsection{Full model FLOPs per step}
Considering $N$ Transformer blocks, the total FLOPs for the blocks is:
\begin{equation}
  F_{\text{blocks}} = N \cdot F_{\text{block}} = N \cdot \left( 24 \cdot L \cdot D^2 + 4 \cdot L^2 \cdot D/H \right).  
\end{equation}

In addition, the embedding and output layers contribute:
\begin{equation}
  F_{\text{others}} = 2\cdot L \cdot D \cdot E + 2\cdot L \cdot (D + 2 \cdot E) \cdot V
\end{equation}

The total FLOPs per step is then:
\begin{equation}
  F_{\text{step}} = F_{\text{blocks}} + F_{\text{others}}.
\end{equation}

\subsection{Total FLOPs for $T$ steps}
For $T$ denoising steps, the total FLOPs is:
\begin{align} \nonumber
F_{\text{DLM}} =&  T \cdot F_{\text{step}} \\ \nonumber 
=& T \cdot \left( N \cdot \left( 24 \cdot L \cdot D^2 + 4 \cdot L^2 \cdot D/H \right) \right.\\ 	
&\left. + 2\cdot L \cdot D \cdot E + 2\cdot L \cdot (D + 4 \cdot E) \cdot V \right)   
\end{align}

Obviously, the square term of L occupies an absolute dominant position in the amount of calculation.
So the time complexity can be written as:
\begin{equation}
T_{\text{DLM}} = \mathcal{O} \left( L^2 \cdot T \right)
\end{equation}

It can be seen that the final time complexity is proportional to the number of denoising steps T and proportional to the square of the sequence length L. For DLMs, \( T \) is a hyperparameter that can be less than the number of tokens generated. By adjusting \( T \), multiple tokens can be generated simultaneously, balancing quality and efficiency. 

\paragraph{Parallel-time complexity.}  
All matrix multiplications inside self-attention and MLP are parallelizable over the sequence dimension~$L$, since the DLM processes the \emph{entire sequence simultaneously} in forward passes.
Assuming ideal hardware that provides $O(L)$ parallel lanes, the \emph{wall-clock latency} per denoising step collapses to the longest dependency path, yielding
\begin{equation}
T_{\text{DLM}}^{\text{step}} = O\!\left(\frac{L D^2}{L} + \frac{L^2 D/H}{L}\right)=O(D^2 + L D/H).
\end{equation}
Multiplying by the number of denoising steps~$T$, the \emph{end-to-end parallel-time complexity} is therefore
\begin{equation}
T_{\text{DLM}}^{\text{par}} = O(T\cdot L).
\end{equation}
Thus, under sufficient parallelism, the latency is \emph{linear in}~$T$ and \emph{linear in}~$L$, while the total FLOPs still scale as~$O(L^{2}T)$.

\section{Detailed FLOPs calculation for autoregressive LLM}
\label{appendix:llm-flops}

In this section, we provide a detailed derivation of the forward inference computational cost (in FLOPs) for decoder-only autoregressive large language models. The analysis includes key components such as attention, feedforward layers, and token embeddings. 

\subsection{Definitions and notations}
We define the following symbols:
\begin{itemize}[itemsep=0pt, topsep=0pt]
    \item $L$: Sequence length.
    \item $D$: Model dimension.
    \item $H$: Number of attention heads.
    \item $N$: Number of Transformer blocks.
    \item $V$: Vocabulary size.
    \item $E$: Embedding dimension (equal to $D$).
\end{itemize}

\subsection{FLOPs of each module}

\textbf{Embedding Layer}
\begin{itemize}[itemsep=0pt, topsep=0pt]
    \item Token embedding: $F_{\text{emb}} =  2\cdot L \cdot D$.
    \item Output projection: $F_{\text{out}} = 2\cdot L \cdot D \cdot V$.
\end{itemize}
Thus, the total FLOPs for the embedding layer is:
\begin{equation}
    F_{\text{embedding}} = 2 \cdot L \cdot D + 2 \cdot L \cdot D \cdot V.
\end{equation}

\textbf{Multi-Head Self-Attention Mechanism (MHSA)}
For a single Transformer block:
\begin{itemize}[itemsep=0pt, topsep=0pt]
    \item Linear projections for $Q$, $K$, and $V$: $6 \cdot L \cdot D^2$.(Consider KV-cache: $6\cdot D^2$)
    \item Attention score computation: $F_{\text{attn\_scores}} = 4 \cdot L^2 \cdot \frac{D}{H}$.(Consider KV-cache: $4 \cdot L \cdot \frac{D}{H}$)
    \item Output projection: $F_{\text{attn\_out}} =2\cdot L \cdot D^2$.(Consider KV-cache: $2\cdot D^2$)
\end{itemize}
The total FLOPs for self-attention is:
\begin{equation}
\begin{split}
F_{\text{SA}} &= 8 \cdot L \cdot D^2 + 4 \cdot L^2 \cdot \frac{D}{H} \\
&\quad (\text{Consider KV-cache:  } 8 \cdot D^2 + 4 \cdot L \cdot \tfrac{D}{H}).
\end{split}
\end{equation}

\textbf{Feedforward Network (FFN)}
The FFN consists of two linear transformations:
\begin{itemize}[itemsep=0pt, topsep=0pt]
    \item Up projection: $F_{\text{ffn\_up}} = 8 \cdot L \cdot D^2$.(\text{Consider KV-cache: } $8 \cdot D^2$).
    \item Down projection: $F_{\text{ffn\_down}} = 8 \cdot L \cdot D^2$.(\text{Consider KV-cache: } $ 8 \cdot D^2$).
\end{itemize}
Thus, the total FLOPs for the FFN is:
\begin{equation}
    F_{\text{FFN}} = 16 \cdot L \cdot D^2.(\text{Consider KV-cache: }  16 \cdot D^2)
\end{equation}

\textbf{Transformer Block}
The total FLOPs for a single Transformer block is:
\begin{align}
F_{\text{block}}
&= F_{\text{SA}} + F_{\text{FFN}} \notag\\
&= 24 \cdot L \cdot D^{2} + 4 \cdot L^{2} \cdot \frac{D}{H}.
\label{eq:block}
\\[4pt]
&\hphantom{=}\ \text{(Consider KV-cache: } 24 \cdot D^{2} + 4 \cdot L \cdot \tfrac{D}{H}\text{)}
\end{align}

So for $N$ Transformer blocks, the total FLOPs are:
\begin{align}
F_{\text{blocks}}
&= N \bigl(24\cdot L\cdot D^{2}\bigr)
   + N \bigl(4\cdot L^{2}\cdot \tfrac{D}{H}\bigr), \notag\\[2pt]
&\quad \text{(Consider KV-cache:} \notag\\
&\quad F^{\prime}_{\text{blocks}}
  = N \bigl(24\cdot D^{2}\bigr)
    + N \bigl(4\cdot L \cdot \tfrac{D}{H}\bigr).\text{)}
\end{align}

\subsection{Total generation FLOPs}
Since each token in the autoregressive language model is generated sequentially, each step requires recomputing self-attention and feedforward layers. The sequence length at each step increases from $L_{\text{in}}$ to $L_{\text{in}} + L_{\text{out}}$. Thus, we approximate the total FLOPs for generation as:

\begin{gather}
F_{\text{total}}
= F_{\text{embedding}} \notag \\
+ L_{\text{out}} \cdot F_{\text{blocks}}\Bigr|_{L \to L_{\text{in}} + L_{\text{out}}} \notag \\
= 2 \cdot L_{\text{in}} \cdot D
+ 2 \cdot L_{\text{out}} \cdot D \cdot V \notag \\
+ 24 \cdot N \cdot L_{\text{out}} \cdot (L_{\text{in}} + L_{\text{out}}) \cdot D^{2} \notag \\
+ 4 \cdot N \cdot L_{\text{out}} \cdot (L_{\text{in}} + L_{\text{out}})^{2} \cdot \frac{D}{H}.
\label{eq:Ftotal-full}
\\[6pt]
\text{(Consider KV-cache: } \quad \notag \\
F_{\text{total}}
= F_{\text{embedding}}
+ L_{\text{out}} \cdot F'_{\text{blocks}}\Bigr|_{L \to L_{\text{in}} + L_{\text{out}}} \notag \\
+ F_{\text{blocks}}\Bigr|_{L \to L_{\text{in}}} \notag \\
= 2 \cdot L_{\text{in}} \cdot D
+ 2 \cdot L_{\text{out}} \cdot D \cdot V \notag \\
+ 24 \cdot N \cdot L_{\text{out}} \cdot D^{2} \notag \\
+ 4 \cdot N \cdot L_{\text{out}} \cdot (L_{\text{in}} + L_{\text{out}}) \cdot \frac{D}{H} \notag \\
+ 24 \cdot N \cdot L_{\text{in}} \cdot D^{2} \notag \\
+ 4 \cdot N \cdot L_{\text{in}}^{2} \cdot \frac{D}{H}. \text{)}
\label{eq:Ftotal-kv}
\end{gather}

Expanding terms:

\begin{gather}
F_{\text{total}}
= 2 \cdot L_{\text{in}} \cdot D
+ 2 \cdot L_{\text{out}} \cdot D \cdot V \notag \\
\quad + 24 \cdot N \cdot L_{\text{out}} \cdot L_{\text{in}} \cdot D^{2} \notag \\
\quad + 24 \cdot N \cdot L_{\text{out}}^{2} \cdot D^{2} \notag \\
\quad + 4 \cdot N \cdot L_{\text{out}} \cdot L_{\text{in}}^{2} \cdot \frac{D}{H} \notag \\
\quad + 8 \cdot N \cdot L_{\text{out}} \cdot L_{\text{in}} \cdot L_{\text{out}} \cdot \frac{D}{H} \notag \\
\quad + 4 \cdot N \cdot L_{\text{out}}^{3} \cdot \frac{D}{H}.
\label{eq:Ftotal-full}
\\[6pt]
\text{(Consider KV-cache:} \quad \notag \\
F_{\text{total}}
= \bigl(2 \cdot D \cdot V + 24 \cdot N \cdot D^{2}\bigr) \cdot L_{\text{out}} \notag \\
\quad + \bigl(2 + 24 \cdot N \cdot D\bigr) \cdot D \cdot L_{\text{in}} \notag \\
\quad + 4 \cdot N \cdot \frac{D}{H} \cdot L_{\text{in}}^{2} \notag \\
\quad + 4 \cdot N \cdot \frac{D}{H} \cdot L_{\text{out}}^{2} \notag \\
\quad + 4 \cdot N \cdot \frac{D}{H} \cdot L_{\text{in}} \cdot L_{\text{out}}.\text{)}
\label{eq:Ftotal-kv}
\end{gather}

\textbf{Asymptotic Complexity:}
\begin{equation}
F_{\text{total}} = \mathcal{O} \left( L_{\text{out}}^3 \right)
(\text{Consider KV-cache: }F_{\text{total}} = \mathcal{O} \left( L_{\text{out}}^2 \right)).
\end{equation}

\paragraph{Sequential-time complexity.}
Unlike diffusion language models, autoregressive language models generate tokens one by one; each new token depends on all previously generated tokens. Consequently, self-attention and MLP layers cannot be parallelized across the sequence dimension~$L_{\text{out}}$, even under ideal hardware.

\subsection{Batch generation comparison}
\label{appendix:timecomplexity}

Assuming the same model architecture, computational resource, parameter sizes, and input/output lengths($L_{in}+L_{out}=L$), the difference in GPU memory occupied by DLM and LLM in the inference process lies in the activation values and KV-cache. For LLM, the number of parameters in the KV-cache can be expressed as:
\begin{equation}
Memory_{KV_{cache}}=2\cdot N \cdot K_{LLM} \cdot D\cdot (L_{in}+L_{out}).
\end{equation}

As for activation values, only the most numerous parts need to be counted:
\begin{itemize}[itemsep=0pt, topsep=0pt]
    \item MHSA: $Memory_{MHSA}=K_{LLM}\cdot L_{in}^2 \cdot H$.
    \item FFN: $Memory_{FFN} = 4\cdot K_{LLM} \cdot L_{in} \cdot D $.
\end{itemize}
For DLM, there is no KV-cache, only the activation values need to be considered: 
\begin{itemize}[itemsep=0pt, topsep=0pt]
    \item MHSA: $Memory_{MHSA}=K_{DLM}\cdot L^2 \cdot H$.
    \item FFN: $Memory_{FFN} = 4\cdot K_{DLM} \cdot L \cdot D $.
\end{itemize}
If $4 \cdot D \geq L \cdot H$ (typically, and in line with our experimental setups): 
\begin{equation}
K_{DLM}/K_{LLM}=\frac{2L_{\text{in}} + N(L_{\text{in}} + L_{\text{out}})}{2(L_{\text{in}} + L_{\text{out}})}.
\end{equation}
Substituting \( N = 32 \):
\begin{equation}
K_{DLM}/K_{LLM}> 16 .
\end{equation}
If $L \cdot H> 4 \cdot D \geq L_{in} \cdot H$: 
\begin{equation}
K_{DLM}/K_{LLM}=\frac{4L_{\text{in}} \cdot D+ 2\cdot D\cdot N\cdot (L_{\text{in}} + L_{\text{out}})}{H\cdot(L_{\text{in}} + L_{\text{out}})^2}.
\end{equation}
Substituting \( H = 32 \),\( N = 32 \):
\begin{equation}
K_{DLM}/K_{LLM}> 2\cdot D/L .
\end{equation}
If $L_{in} \cdot H> 4 \cdot D$ : 
\begin{equation}
K_{DLM}/K_{LLM}=\frac{L_{\text{in}}^2 \cdot H+ 2\cdot D\cdot N\cdot (L_{\text{in}} + L_{\text{out}})}{H\cdot(L_{\text{in}} + L_{\text{out}})^2}.
\end{equation}
Substituting \( H = 32 \),\( N = 32 \):
\begin{equation}
K_{DLM}/K_{LLM}> 2\cdot D/L .
\end{equation}


    



\section{Error analysis in diffusion language models}
\label{appendix:information}
We extend the information-theoretic approach in Gan et al.~\cite{gan2025rethinking} to analyze diffusion language models. Diffusion models generate a sequence $X_0$ of length $L$ by iteratively denoising an initial noise sequence $X_T$ over $T$ steps: $X_T \rightarrow X_{T-1} \rightarrow \dots \rightarrow X_1 \rightarrow X_0$. A common practice is to predict tokens in parallel at each step for efficiency.

\subsection{Information loss from independence assumption}

Let $X_t = (x_1^t, \dots, x_L^t)$ be the sequence at denoising step $t$. Consider the transition from $X_{t-1}$ to $X_t$. The true conditional probability is $p(X_t | X_{t-1})$, and the associated conditional entropy is $\Ent(X_t | X_{t-1})$.

Many diffusion models approximate the reverse process by assuming conditional independence of tokens given the previous state:
\begin{equation} p_\theta(X_t | X_{t-1}) = \prod_{i=1}^{L} p_\theta(x_i^t | X_{t-1}) \end{equation}
Under this assumption, the conditional entropy calculated by the model is:
\begin{equation} \Ent_{\text{indep}}(X_t | X_{t-1}) = \sum_{i=1}^{L} \Ent(x_i^t | X_{t-1}) \end{equation}
where the entropy $\Ent(x_i^t | X_{t-1})$ is computed based on the marginal $p_\theta(x_i^t | X_{t-1})$.

As derived from the chain rule of entropy and the property that conditioning reduces entropy:
\begin{equation}
\begin{split}
\Ent(X_t | X_{t-1}) &= \sum_{i=1}^{L} \Ent(x_i^t | x_1^t, \dots, x_{i-1}^t, X_{t-1}) \\
&\le \sum_{i=1}^{L} \Ent(x_i^t | X_{t-1}) = \Ent_{\text{indep}}(X_t | X_{t-1})
\end{split}
\end{equation}
Thus, the independence assumption leads to an overestimation of the conditional entropy:
\begin{equation} \Ent(X_t | X_{t-1}) \le \Ent_{\text{indep}}(X_t | X_{t-1}) \end{equation}
This implies an underestimation of the mutual information between consecutive steps:
\begin{equation} \MI(X_t; X_{t-1}) = \Ent(X_t) - \Ent(X_t | X_{t-1}) \end{equation}
\begin{equation} \MI_{\text{indep}}(X_t; X_{t-1}) = \Ent(X_t) - \Ent_{\text{indep}}(X_t | X_{t-1}) \end{equation}
\begin{equation} \implies \MI_{\text{indep}}(X_t; X_{t-1}) \le \MI(X_t; X_{t-1}) \end{equation}

\begin{definition}[Per-Step information loss in diffusion]
The information loss at denoising step $t$ due to the independence assumption is the difference between the model's estimated conditional entropy and the true conditional entropy:
\begin{equation} \Delta \Ent_t = \Ent_{\text{indep}}(X_t | X_{t-1}) - \Ent(X_t | X_{t-1}) \ge 0 \end{equation}
Alternatively, it's the difference in mutual information:
\begin{equation} \Delta \MI_t = \MI(X_t; X_{t-1}) - \MI_{\text{indep}}(X_t; X_{t-1}) \ge 0 \end{equation}
This loss $\Delta \MI_t$ quantifies the information about the dependencies between tokens $x_i^t$ within step $t$ (given $X_{t-1}$) that is discarded by the model. This loss is expected to increase with sequence length $L$.
\end{definition}

\subsection{Cumulative loss and final error probability}

Let $X^*$ be the ground truth sequence the model aims to generate. The difference in mutual information hinders the model's ability to maximize the mutual information between the final output $X_0$ and the target $X^*$, denoted $\MI(X^*; X_0)$.

Let $\MI_{\text{ideal}}(X^*; X_0)$ be the mutual information achievable by an ideal T-step diffusion process (without the independence assumption), and $\MI_{\text{indep}}(X^*; X_0)$ be the mutual information achieved by the model using the independence assumption. The total accumulated information loss is:
\begin{equation} \text{TotalLoss}(T, L) = \MI_{\text{ideal}}(X^*; X_0) - \MI_{\text{indep}}(X^*; X_0) \ge 0 \end{equation}
This loss reflects the cumulative impact of $\Delta \MI_t$ for $t = T, \dots, 1$.

Let $E_{Diff}(T, L) = \Prob(X_0 \ne X^*)$ be the final error probability. Using Fano's inequality on the pair $(X^*, X_0)$:
\begin{equation} \Ent(X^* | X_0) \le \Ent(E_{Diff}(T, L)) + E_{Diff}(T, L) \log(|\mathcal{X}| - 1) \end{equation}
where $\mathcal{X}$ is the space of possible sequences.

The conditional entropy for the actual process is:
\begin{equation}
\begin{split}
    \Ent_{\text{indep}}(X^* | X_0) = \Ent(X^*) - \MI_{\text{indep}}(X^*; X_0) \end{split}\end{equation}
Substituting the definition of TotalLoss:
\begin{equation}\begin{split} \Ent_{\text{indep}}(X^* | X_0) = \Ent(X^*) - (\MI_{\text{ideal}}(X^*; X_0) \\ - \text{TotalLoss}(T, L)) \end{split}\end{equation}
\begin{equation} \begin{split}\Ent_{\text{indep}}(X^* | X_0) = (\Ent(X^*) - \MI_{\text{ideal}}(X^*; X_0)) \\ + \text{TotalLoss}(T, L) \end{split}\end{equation}
\begin{equation}\begin{split} \Ent_{\text{indep}}(X^* | X_0) = \Ent_{\text{ideal}}(X^* | X_0) + \text{TotalLoss}(T, L)\end{split} \end{equation}
where $\Ent_{\text{ideal}}(X^* | X_0)$ is the conditional entropy of the ideal process.

Plugging this back into the Fano inequality bound:
\begin{equation}\begin{split}
\Ent_{\text{ideal}}(X^* | X_0) + \text{TotalLoss}(T, L) \le \Ent(E_{Diff}(T, L)) \\+ E_{Diff}(T, L) \log(|\mathcal{X}| - 1) \end{split} \end{equation}

\textbf{Implication:} This inequality shows that the final error probability $E_{Diff}(T, L)$ is lower-bounded by a term that grows with the total accumulated information loss $\text{TotalLoss}(T, L)$ resulting from the independence assumption across the $T$ steps.

\subsection{Dependence on L and T}

\textbf{Sequence length ($L$)}: Since the per-step loss $\Delta \MI_t$ likely increases with $L$ (more dependencies to ignore), the cumulative loss $\text{TotalLoss}(T, L)$ also increases with $L$. Therefore, the final error $E_{Diff}(T, L)$ is expected to increase with $L$.

\textbf{Number of steps ($T$)}: The dependence is complex. Increasing $T$ allows more denoising (information recovery about $X^*$, reducing $\Ent_{\text{ideal}}(X^* | X_0)$). Empirically, $E_{Diff}(T, L)$ decreases with $T$ up to a point, indicating the denoising benefit usually dominates. However, the minimum error is limited by $\text{TotalLoss}(T, L)$.


\section{Discussions}\label{discussion}

\subsection{Implementation details}\label{details}
\begin{table}[H]

\centering
\begin{tabular}{@{}ccc@{}}

\toprule
Module                    & Element              & Detail                                  \\ \midrule
\multirow{9}{*}{System}   & OS                   & Ubuntu 20.04.6 LTS                          \\
                          & CUDA                 & 12.4                                    \\
                          & Python               & 3.10.16                                  \\
                          & Pytorch              & 2.6.0 \\
                          & deepspeed         & 0.16.8+ee492c30    \\
                          & accelerate         & 1.4.0    \\
                          & peft         & 0.15.1    \\
                          & Device               & 1*NVIDIA A100 80G                       \\ \midrule
\multirow{5}{*}{Llama3-70B}       
                                                      & Framework            & vLLM(0.9.2)                        \\
                          & Tensor Parallel      & 2                           \\
                          & Data Type            & FP16                        \\
                          & Max Length           & 8192                        \\
                          & GPU Utilization      & 0.93                        \\
                            \midrule
\multirow{8}{*}{SFT}       
                            & Mode  & Lora \\
                            & Batch size      & 2  \\
                          & Number of epochs     & 1                                      \\
\multicolumn{1}{l}{}      & Max token length     & 4096                                     \\                                        & LoRA rank           & 128                                         \\
                          & Optimzer             & AdamW                                    \\
                          & Learning rate        & 0.00001                                 \\

                           \bottomrule
\end{tabular}

\vspace{3mm}
\caption{\textbf{Detailed Experimental Settings}}
\label{tbl:details}
\end{table}

Here we provide detailed experimental settings
to facilitate the reproducibility of our results in Table~\ref{tbl:details}. All experiments were run three times except for the scaling of proposals for one time.

\subsection{Code of ethics}
\label{code of ethics}
In this paper, we use open-source models, which involve no problem regarding privacy and copyright. We use open-source datasets and self-constructed datasets, which involve no problem regarding privacy and copyright. Except for Game-of-24, we all use open-source datasets. For the Game-of-24 , we exhaustively enumerated num-
ber combinations and employed a backtracking approach to
solve it. We considered all non-repetitive, diverse solutions,
resulting in a dataset of 168,046 24-point problems, with
the maximum value for the four numbers limited to 30 to make it more challenging. We have cited all open-source resources.
Our project code has also been released and is available through the following anonymous link: \url{https://anonymous.4open.science/r/Diffuse-Thinking-EC60}. The self-constructed data used in this paper is also released at the provided anonymous link.

\subsection{Broader impacts}
\label{broader impacts}
This research enhances complex reasoning efficiency by combining diffusion language models (DLMs) with large language models (LLMs), making problem-solving faster and more scalable. The work also advances AI research by exploring collaborative frameworks, inspiring new methodologies. These improvements can be applied to real-world issues in various fields, leading to more efficient solutions and benefiting society through technological advancements.

\section{Case study}\label{case study}
To better understand the mechanism of our framework, we examined a representative case study on the game of 24.

We have four numbers: 1, 14, 16, and 25. We provide DLM with prompt:
\definecolor{Periwinkle}{RGB}{255, 228, 225}  
\definecolor{Yellow}{RGB}{250, 241, 195}  
\begin{tcolorbox}[notitle, sharp corners, breakable, 
     colframe=Periwinkle, colback=white, 
     boxrule=3pt, boxsep=0.5pt, enhanced, 
     shadow={3pt}{-3pt}{0pt}{opacity=0.3},title={Prompt}]
     \footnotesize
     {\fontfamily{pcr}\selectfont
     \spaceskip=0pt plus 0pt minus 0pt 
\begin{lstlisting}[breaklines=true,showstringspaces=false]
Here is a task for you: use these numbers <<1,14,16,25>>  to obtain 24 through the basic operation of (+- */). Each number can only be used once and must be used.
Please output the next possible operation directly for only one line, in the format of: Equation (remaining numbers)
\end{lstlisting}
}
\end{tcolorbox}

DLM generates the following text in parallel:
\begin{tcolorbox}[notitle, sharp corners, breakable, 
     colframe=Yellow, colback=white, 
     boxrule=3pt, boxsep=0.5pt, enhanced, 
     shadow={3pt}{-3pt}{0pt}{opacity=0.3},
     title={Response}]
     \footnotesize
     {\fontfamily{pcr}\selectfont
     \spaceskip=0pt plus 0pt minus 0pt 
\begin{lstlisting}[breaklines=true,showstringspaces=false]
'16-1=15 (14,25,15)', '16-1=15 (14,25,15)', '1*25=25 (15,16,25)', '16*6=15 (14,25,15)', '16-1=15 (14,25,15)', '1*15=15 (25,14,15)', '1*25=25 (25,14,25)', '14+1=15 (16,25,15)', '14+1=15 (16,25,15)', '1+14=15 (16,25,15)', '1*11=15 (14,25,15)', '1+25=16 (25,16,25)', '14+1=15 (25,16,15)', '16+6=17 (14,25,17)', '20-1=19 (14,16,19)', '16-1=15 (25,16,17)', '1*15=15 (25,14,16)', '16-1=15 (14,25,15)', '16+1=17 (14,25,17)', '1*25=16 (26,14,16)'
\end{lstlisting}
}
\end{tcolorbox}

Then we label them and provide LLM with prompt:
\begin{tcolorbox}[notitle, sharp corners, breakable, 
     colframe=Periwinkle, colback=white, 
     boxrule=3pt, boxsep=0.5pt, enhanced, 
     shadow={3pt}{-3pt}{0pt}{opacity=0.3},
     title={Prompt}]
     \footnotesize
     {\fontfamily{pcr}\selectfont
     \spaceskip=0pt plus 0pt minus 0pt 
\begin{lstlisting}[breaklines=true,showstringspaces=false]
You must consider whether the expression calculation in the next thought proposal<<>>is correct,
whether the number on the left side of the equation is in the remaining<<>>, 
whether the number on the right side of the equation is in the left<<>>, and 
whether the number in the left<<>>is only '24' left or more likely to achieve 24 through basic arithmetic operations (+- */).
Here are some candidate solutions for the next step.Their serial numbers are in [].[1]16-1=15 (14,25,15),[2]16-1=15 (14,25,15), [3]1*25=25 (15,16,25), [4]16*6=15 (14,25,15), [5]16-1=15 (14,25,15), [6]1*15=15 (25,14,15), [7]1*25=25 (25,14,25), [8]14+1=15 (16,25,15), [9]14+1=15 (16,25,15), [10]1+14=15 (16,25,15), [11]1*11=15 (14,25,15), [12]1+25=16 (25,16,25), [13]14+1=15 (25,16,15), [14]16+6=17 (14,25,17), [15]20-1=19 (14,16,19), [16]16-1=15 (25,16,17), [17]1*15=15 (25,14,16), [18]16-1=15 (14,25,15), [19]16+1=17 (14,25,17), [20]1*25=16 (26,14,16)
Please choose the best one and tell me the serial number you have chosen.OUTPUT FORMAT:'Reasons:....  [serial  number]'.
\end{lstlisting}
}
\end{tcolorbox}

LLM responses:
\begin{tcolorbox}[notitle, sharp corners, breakable, 
     colframe=Yellow, colback=white, 
     boxrule=3pt, boxsep=0.5pt, enhanced, 
     shadow={3pt}{-3pt}{0pt}{opacity=0.3},
     title={Response}]
     \footnotesize
     {\fontfamily{pcr}\selectfont
     \spaceskip=0pt plus 0pt minus 0pt 
\begin{lstlisting}[breaklines=true,showstringspaces=false]
After analyzing the given options, I found that most of the correct calculations are not achieving the target of getting closer to 24 or ensuring that the numbers involved are or can lead to 24 through basic arithmetic operations.

The correct calculations that lead to correct results but do not necessarily help achieve 24 or are not the focus based on the provided context are numerous. However, to identify the best candidate that fits the criteria of correct calculation and the potential to contribute to reaching 24 or being part of a sequence that could, with further operations, result in 24, I must consider the arithmetic operations' correctness and the numbers involved.

Given the constraints and looking at the patterns and potential for leading to 24, I notice that many of these equations simply demonstrate arithmetic operations without a clear progression towards achieving 24. To select the best option, I would look for operations that either directly involve numbers close to 24 or demonstrate a path that could, with additional operations, lead to 24.

However, since most provided equations do not directly aim at achieving 24 or do not specify the next steps in calculations that would lead to 24, the choice seems to be more about identifying correct arithmetic rather than progression towards a specific goal.

Given this context and looking for an option that seems plausible or directly correct in its arithmetic without the additional context of how it leads to 24 (since such context is not provided in most options), I would choose an equation that is both correct and involves numbers that could potentially lead to interesting further calculations.

Considering the arithmetic correctness and the involvement of numbers that could be part of further calculations leading to 24 (though none directly achieve this), I pick an option that demonstrates a straightforward and correct arithmetic operation.

Reasons: The equation is simple, the arithmetic is correct, and it involves basic addition which could be a starting point for further operations. However, given the constraint and looking for the best representation of arithmetic that could lead to further calculations, [8] stands out for its simplicity and correctness: 14+1=15, which is a straightforward and correct operation, even though it doesn't directly aim for 24. 
[8]
\end{lstlisting}
}
\end{tcolorbox}

We obtain [8] 14+1=15 (16, 25, 15), and then this process will continue to iterate until the problem is solved.

\end{document}